\documentclass{article}

\PassOptionsToPackage{numbers, compress}{natbib}
\usepackage[preprint]{neurips_2025}

\usepackage[utf8]{inputenc} % allow utf-8 input
\usepackage[T1]{fontenc}    % use 8-bit T1 fonts
\usepackage{hyperref}       % hyperlinks
\usepackage{url}            % simple URL typesetting
\usepackage{booktabs}       % professional-quality tables
\usepackage{amsfonts}       % blackboard math symbols
\usepackage{nicefrac}       % compact symbols for 1/2, etc.
\usepackage{microtype}      % microtypography
\usepackage{xcolor}         % colors
\usepackage{graphicx}
\usepackage{rotating}
\usepackage{colortbl}  % For coloring the row background
\usepackage{subcaption}
\usepackage{stfloats}
\usepackage{wrapfig}
\usepackage{amsmath}
\usepackage{array} 
\usepackage{colortbl}
\usepackage{makecell}
\usepackage{multirow}

\title{Constrained Prompt Enhancement for Improving Zero-Shot Generalization of Vision-Language Models}

\author{%
  Xiaojie Yin \\
  Tianjin University \\
  \texttt{xjyin@tju.edu.cn} \\
  \And
  Qilong Wang* \\
  Tianjin University \\
  \texttt{qlwang@tju.edu.cn} \\
  \AND
  Qinghua Hu \\
  Tianjin University \\
  \texttt{huqinghua@tju.edu.cn} \\
}

\begin{document}

\maketitle

\begin{abstract}
Vision-language models (VLMs) pre-trained on web-scale data exhibit promising zero-shot generalization but often suffer from semantic misalignment due to domain gaps between pre-training and downstream tasks. Existing approaches primarily focus on text prompting with class-specific descriptions and visual-text adaptation via aligning cropped image regions with textual descriptions. However, they still face the issues of incomplete textual prompts and noisy visual prompts. In this paper, we propose a novel constrained prompt enhancement (CPE) method to improve visual-textual alignment by constructing comprehensive textual prompts and compact visual prompts from the semantic perspective. Specifically, our approach consists of two key components: Topology-Guided Synonymous Semantic Generation (TGSSG) and Category-Agnostic Discriminative Region Selection (CADRS). Textually, to address the issue of incomplete semantic expression in textual prompts, our TGSSG first generates synonymous semantic set for each category via large language models, and constructs comprehensive textual prompts based on semantic ambiguity entropy and persistent homology analysis. Visually, to mitigate the irrelevant visual noise introduced by random cropping, our CADRS identifies discriminative regions with activation maps outputted by a pre-trained vision model, effectively filtering out noisy regions and generating compact visual prompts. Given the comprehensive set of textual prompts and compact set of visual prompts, we introduce two set-to-set matching strategies based on test-time adaptation (TTA) and optimal transport (OT) to achieve effective visual-textual alignment, and so improve zero-shot generalization of VLMs. Extensive experiments on 10 zero-shot image classification benchmarks, 5 zero-shot out-of-distribution tasks, and 3 zero-shot video action recognition datasets demonstrate that our CPE method clearly outperforms its counterparts, while achieving state-of-the-art performance.
\end{abstract}

\section{Introduction}
\label{sec:intro}

%-------------------------------------------------------------------------
Undergoing extensive pre-training on web-scale image-text pairs~\cite{radford2021learning,schuhmann2021laion,schuhmann2022laion}, vision-language models (VLMs) have demonstrated promising zero-shot generalization ability. VLMs~\cite{radford2021learning,jia2021scaling,zhai2022lit,ge2022dall,ramesh2021zero} are trained to associate images with relevant textual descriptions. However, there often exists a domain gap between pre-training data and that in domain-specific downstream tasks~\cite{mehrabi2021survey,menon2022task,agarwal2021evaluating}, especially the annotation mode of textual concepts. This intuitively leads to semantic misalignment between image and text in the feature space defined by pre-trained VLMs, limiting the zero-shot generalization ability.

\begin{table*}[t]
    \footnotesize
    \renewcommand{\arraystretch}{1.2}
    \centering
    \setlength{\tabcolsep}{2pt}
    \begin{tabular}{>{\raggedright\arraybackslash}p{1.8cm}>{\centering\arraybackslash}p{1.8cm}>{\centering\arraybackslash}p{3.2cm}>{\centering\arraybackslash}p{5cm}>{\centering\arraybackslash}p{1.4cm}}
    \toprule
    Method & Alignment & Visual Prompt & Textual Prompt & \textit{Avg. Acc.} \\
    \midrule
    CLIP~\cite{radford2021learning} & Point-to-Point & - & a photo of a \{class\} & 63.58 \\
    DCLIP~\cite{menon2022visual} & Point-to-Set & - & a photo of a \{class\}, \{description\} & 66.27 \\
    TPS~\cite{sui2024just} & Set-to-Set & Random Aug. & a photo of a \{class\}, \{description\} & 66.95 \\
    AWT~\cite{zhu2024awt} & Set-to-Set & Random Aug. & a photo of a \{class\}, \{description\} & 70.51 \\
    CPE (Ours) & Set-to-Set & \textbf{Region Selection} Aug. & a photo of a \{\textbf{synonym}\}, \{description\} & \textbf{72.56} \\
    \bottomrule
    \end{tabular}
    \caption{\textbf{Comparison of existing CLIP-based zero-shot generalization methods from the visual-textual alignment.} Different from previous works, our CPE improves upon set-to-set alignment by incorporating constraints on visual prompts and textual prompts, while achieving higher average accuracy on 10 zero-shot image classification datasets.}
    \label{tab:comparison}
\end{table*}

\begin{figure}[t]
    \centering
    \begin{minipage}[b]{0.44\textwidth}
        \centering
        \includegraphics[width=\textwidth]{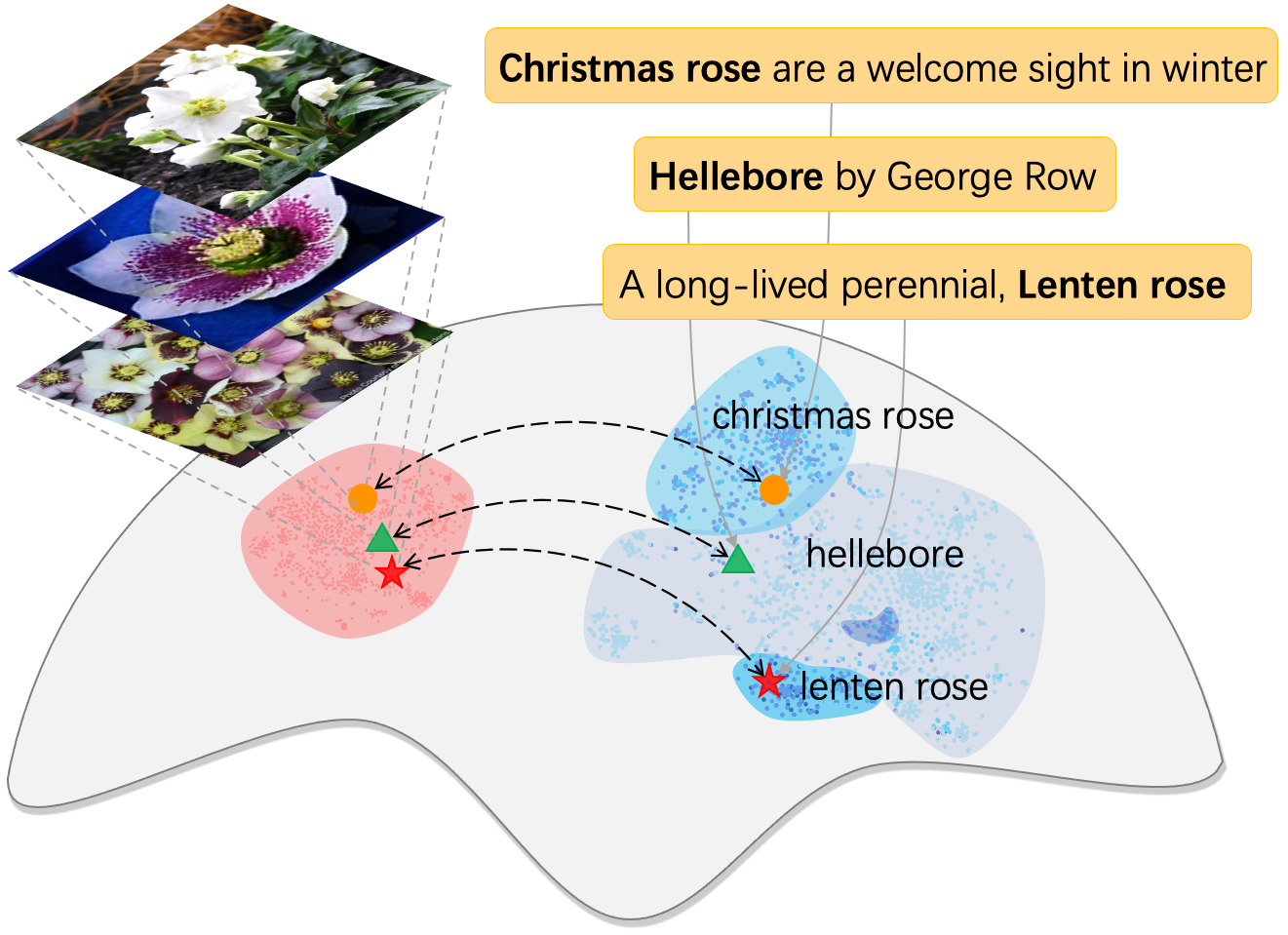}  
        \subcaption{Lexical variation in LAION-400M dataset.}
        \label{fig:laion}
    \end{minipage}
    \hfill
    \begin{minipage}[b]{0.47\textwidth} 
        \centering
        \includegraphics[width=\textwidth]{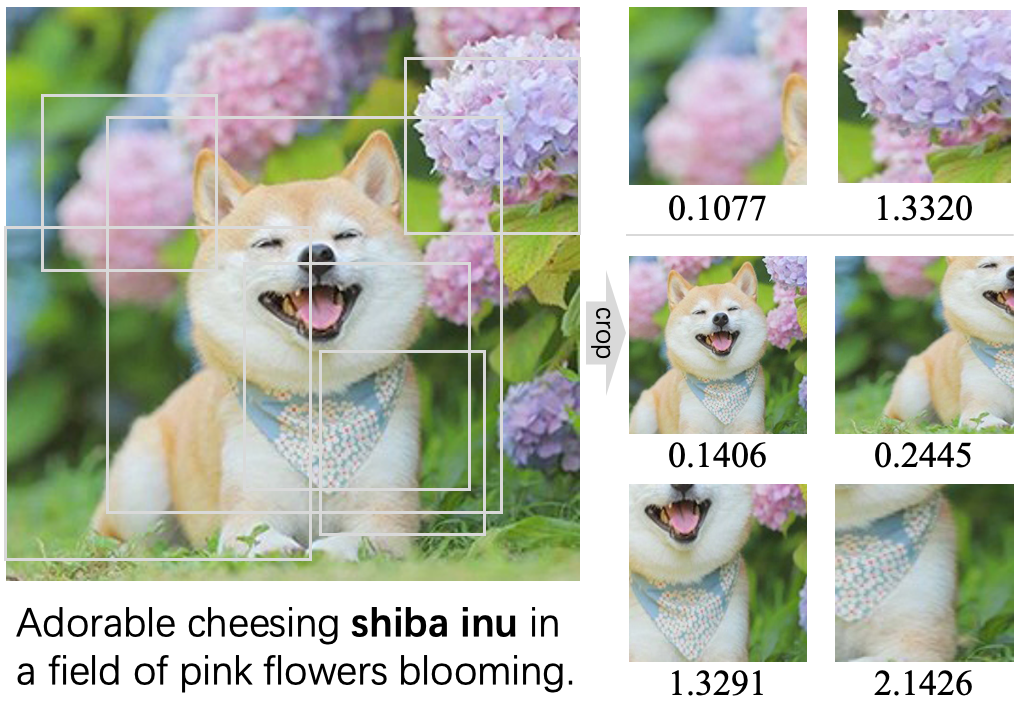}
        \subcaption{Random cropping introduces irrelevant noise.}
        \label{fig:crop}
    \end{minipage}
    \caption{\textbf{(a) Lexical variation in LAION-400M dataset:} We retrieved captions and corresponding images of the concept ``hellebore'' from the web-scale pre-trained dataset LAION-400M~\cite{schuhmann2021laion} and visualized their CLIP embeddings. \textbf{(b) Random cropping introduces irrelevant noise:} We perform random cropping on an image of a dog and present six cropped views along with their matching entropy values with respect to all labels in the Oxford Pets. }
\end{figure}

\begin{figure}[t]
   \centering
   \includegraphics[width=\linewidth]{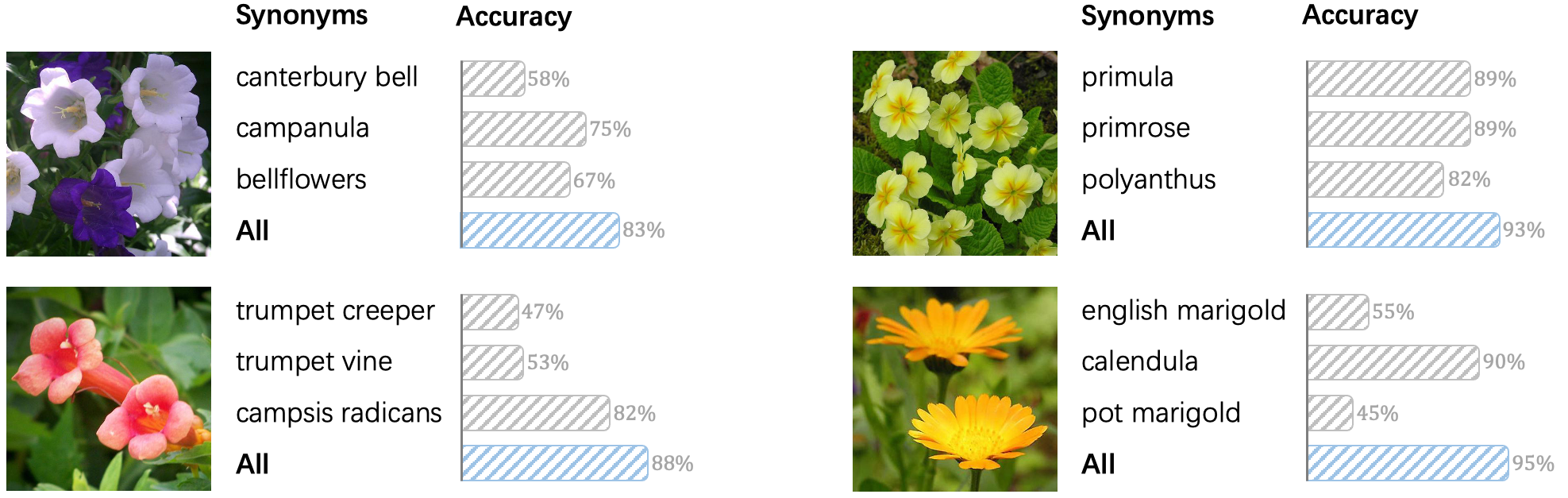}
   \caption{\textbf{Effect of synonyms selection.} We replace the given concept name (e.g., ``canterbury bell'') with its synonyms (e.g., ``campanula'' and ``bellflowers'') in the prompt template, such as ``a photo of a \{concept\}''. Additionally, we employ prompts that include all available synonyms. We demonstrate the impact of synonym selection on classification accuracy for various Oxford Flowers concepts. Using all available synonyms yields superior performance compared to using a single synonym.}
   \label{fig:synonym}
\end{figure}

To address the above issue, several promising solutions have been proposed, which can be roughly categorized into text prompting and visual-text adaptation. Specifically, text prompting, such as DCLIP~\cite{menon2022visual}, focus on employing large language models (LLMs) to generate multiple and more detailed textual descriptions of each category. This strategy leverages the prior knowledge embedded in LLMs, allowing the model to better capture category-specific features and improve alignment with visual content (Row 2 of Table \ref{tab:comparison}). As the generated textual descriptions become more refined and comprehensive, the zero-shot performance of existing methods gradually improves. However, VLMs pre-trained on web-scale datasets inevitably face the challenge of lexical variation~\cite{parashar2024neglected}, as a single concept can be expressed using multiple synonyms. For instance, textual prompts for the concept of ``hellebore'' can include synonyms such as ``hellebore'', ``christmas rose'', and ``lenten rose''. As illustrated in Figure \ref{fig:laion}, text embeddings of synonyms are relatively far away each others, but image embeddings are close to each other in feature space. Besides, as shown in Figure \ref{fig:synonym}, classification accuracy significantly changes along various selections of synonyms on Oxford Flowers, while a comprehensive textual set consistently enhances the model's zero-shot recognition accuracy significantly. Therefore, these synonyms collectively contribute to semantic representation of one textual concept, and incomplete textual semantics will limit the zero-shot performance of VLMs. 

Building upon text prompting, visual-text adaptation aims to enhance visual inputs. In contrast to aligning textual descriptions with the whole image, visual-text adaptation (such as TPS~\cite{sui2024just} and AWT~\cite{zhu2024awt}) randomly crop the image into multiple regions and focus on aligning specific areas with textual descriptions, thereby reducing the semantic bias that arises from global alignment (Rows 3-4 of Table \ref{tab:comparison}). However, random cropping introduces irrelevant noise to the target. As shown in Figure \ref{fig:crop}, an image of a dog might contain irrelevant and noisy elements such as a ``flower'' or ``grass''. Previous works~\cite{sui2024just,zhu2024awt,li2024visual} have been studied to eliminate noisy regions with high image-text matching entropy, but this approach may not always be practical. An shown in Figure \ref{fig:crop}, two background regions (i.e., ``flowers'' (misclassified as ``Abyssinian'' and ``German Shorthair'') are irrelevant to the target ``Shiba Inu'', but they have lower matching entropy than some ones of the cropped regions containing the target, making them difficult to be identified as noise. In conclusion, although existing methods have shown promising results, they still face challenges of incomplete textual prompts and noisy visual prompts.

\begin{figure}[t]
   \centering
   \includegraphics[width=\linewidth]{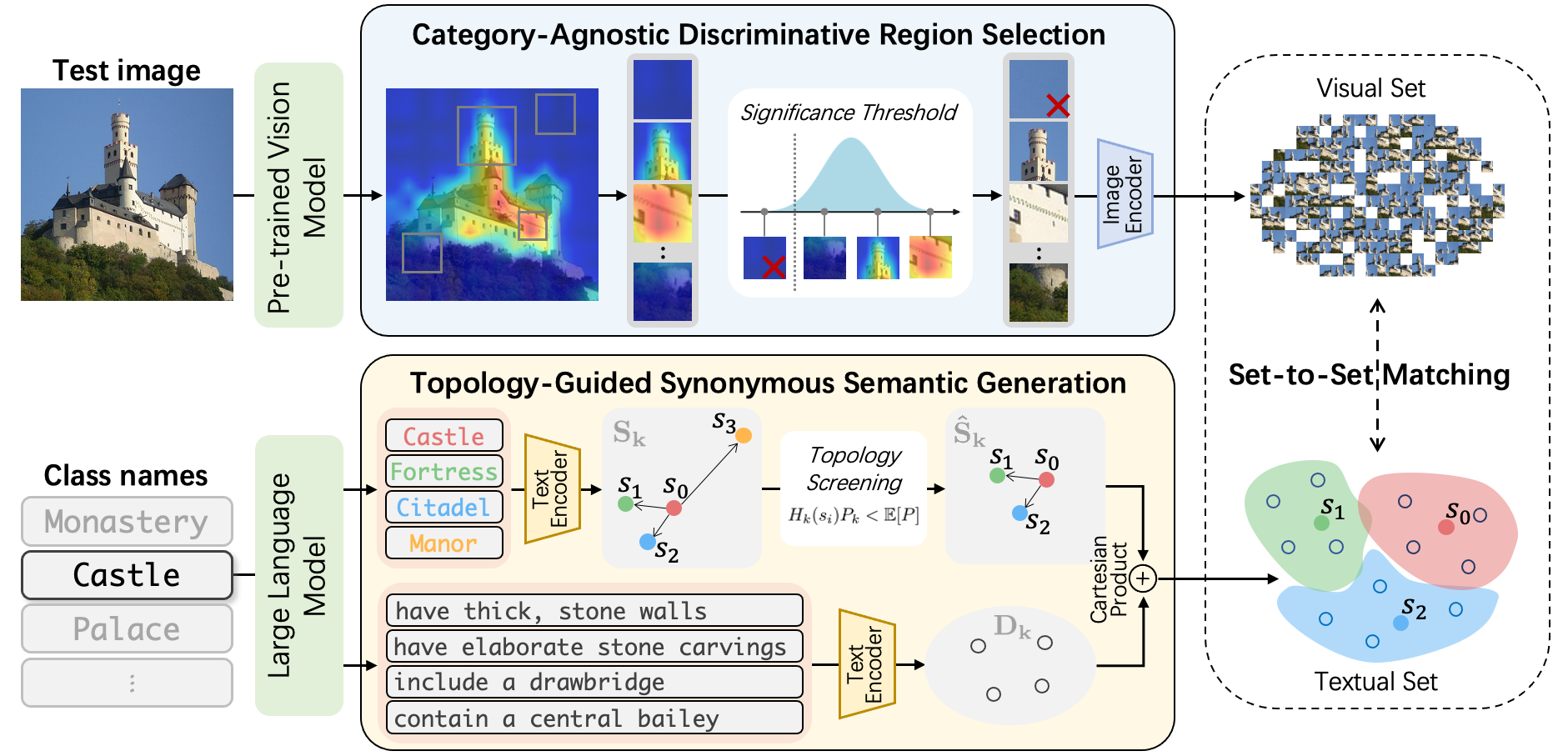}

   \caption{\textbf{Overall architecture of CPE.} For each candidate class, synonyms and descriptions are generated using LLMs and processed by the Topology-Guided Synonymous Semantic Generation to form a comprehensive textual set. A test image is passed through the pre-trained vision model to obtain an activation map, which is then processed by the Category-Agnostic Discriminative Region Selection to create a compact visual set. The Set-to-Set Visual-Textual Matching evaluates the predicted probability of the visual set aligning with each class's textual set.}
   \label{fig:main}
\end{figure}

To address these challenges, we propose a novel constrained prompt enhancement (CPE) method, whose core idea is to perform a set-to-set matching between comprehensive textual prompts and compact visual prompts for improving visual-textual alignment, thereby enhancing zero-shot generalization of VLMs. To this end, our CPE method consists of two key components: Topology-Guided Synonymous Semantic Generation (TGSSG) and Category-Agnostic Discriminative Region Selection (CADRS). Specifically, our \textbf{TGSSG} first generates synonyms and descriptions for class labels by leveraging some advanced large language models (e.g., Claude~\cite{anthropic2024claude} and GPT-4~\cite{achiam2023gpt}), while introducing semantic ambiguity entropy and persistent homology analysis to filter out irrelevant candidates. As such, our TGSSG constructs comprehensive textual prompts, addressing the issue that existing textual prompts cannot fully capture the intended semantics. On another hand, our \textbf{CADRS} feeds images into a pre-trained vision model (e.g., DINO~\cite{caron2021emerging}), and identifies the discriminative regions while filtering  the irrelevant noise by performing a statistical threshold of the outputted activation maps. As such, our CADRS constructs compact visual prompts, mitigating the irrelevant and noisy views introduced by random cropping. Based on the generated comprehensive set of textual prompts and compact set of visual prompts, we introduce two set-to-set matching strategies based on test-time adaptation (TTA) and optimal transport (OT) to perform effective visual-textual alignment, namely CPE-TTA and CPE-OT. Specifically,  CPE-TTA dynamically adapts text embeddings to visual content via entropy minimization during test time, and CPE-OT aligns visual and textual sets by solving a mass transportation problem with entropy-regularized cost matrices. The overview of our method is illustrated in Figure~\ref{fig:main}. To evaluate the effectiveness of our CPE method, experiments are conducted on ten zero-shot image classification tasks (i.e.,  Oxford Flowers~\cite{nilsback2008automated}, DTD~\cite{cimpoi2014describing}, Oxford Pets~\cite{parkhi2012cats}, Stanford Cars~\cite{krause20133d}, UCF101~\cite{soomro2012ucf101}, Caltech101~\cite{fei2004learning}, Food101~\cite{bossard2014food}, SUN397~\cite{xiao2010sun}, FGVC Aircraft~\cite{maji2013fine}, and EuroSAT~\cite{helber2019eurosat}),  five zero-shot out-of-distribution generalization tasks (e.g., ImageNet~\cite{deng2009imagenet}, ImageNet-A~\cite{hendrycks2021natural}, ImageNet-V2~\cite{recht2019imagenet}, ImageNet-R~\cite{hendrycks2021many}, and ImageNet-Sketch~\cite{wang2019learning}), and three zero-shot video action recognition tasks (e.g., UCF101~\cite{soomro2012ucf101}, HMDB51~\cite{kuehne2011hmdb}, and Kinetics-600~\cite{carreira2018short}). The contributions of this work are summarized as follows:

(i) In this paper, we propose a novel constrained prompt enhancement (CPE) method, whose core idea is to perform a set-to-set matching between comprehensive textual prompts and compact visual prompts for improving visual-textual alignment of VLMs.

(ii) To this end, our CPE presents Topology-Guided Synonymous Semantic Generation (TGSSG) and Category-Agnostic Discriminative Region Selection (CADRS) to generate comprehensive textual prompts and compact visual prompts, respectively. Subsequently, two set-to-set matching strategies based on test-time adaptation (TTA) and optimal transport (OT) are introduced to align the enhanced textual prompts and enhanced visual prompts. 

(iii) Extensive experiments demonstrate our CPE method clearly outperforms its counterparts, while achieving state-of-the-art performance.

\section{Related Work}
\label{sec:related}

%-------------------------------------------------------------------------
\paragraph{Vision-language models.}
Leveraging extensive pre-training on web-scale image-text pairs, vision-language models (VLMs) such as CLIP~\cite{radford2021learning}, ALIGN~\cite{jia2021scaling}, SigLIP~\cite{zhai2023sigmoid}, and EVA-CLIP~\cite{sun2023eva} embed both texts and images into a shared feature space. This facilitates the proximity of semantically similar inputs, resulting in exceptional performance across a wide range of open-ended tasks, including image classification~\cite{li2022language,zhou2022learning,zhou2022conditional}, object detection~\cite{feng2022towards,gu2021zero,du2022learning}, and video action recognition~\cite{wang2021actionclip,weng2023open,huang2024froster}. However, pre-trained VLMs often exhibit biases related to gender, race, and geography~\cite{parashar2024neglected,mehrabi2021survey}, result in skewed predictions in downstream tasks. This study aims to mitigate these biases by incorporating comprehensive textual prompts and compact visual prompts within VLMs.

\paragraph{Textual prompting in VLMs.}
CLIP~\cite{radford2021learning} suggests that placing a given concept name in human-engineered prompt templates, such as ``a photo of a \{class\}'' or ``a demonstration of a \{class\}'', often improves zero-shot recognition performance, and highlights that the choice of prompt significantly influences the performance of downstream tasks~\cite{zhou2022learning,zhou2022conditional}. DCLIP~\cite{menon2022visual} and CuPL~\cite{pratt2023does} leverage the knowledge contained in LLMs, such GPT-3~\cite{brown2020language} to automatically generate class-specific descriptions, enriching the semantic content of textual prompts and leading to more effective results. WaffleCLIP~\cite{roth2023waffling} and MPVR~\cite{mirza2024meta} further confirm that combining concept names with descriptions in prompts (e.g., ``a photo of a \{class\}, \{description\}'') significantly reduces semantic bias and achieves improved performance in downstream tasks. Recent study~\cite{parashar2024neglected} shows that replacing the given class name with its most common name in the prompt can improve zero-shot recognition for fine-grained species. In contrast, this study advocates that multiple synonyms collectively contribute to semantic representation of one textual concept. 

\paragraph{Visual-text adaptation in VLMs.}
In contrast to focusing solely on text augmentation, visual-text adaptation aims to improve alignment by processing both visual and textual inputs. WCA~\cite{li2024visual} suggests that more detailed textual descriptions may align more accurately with specific areas of an image, enabling better cross-alignment of finer descriptions with local visual regions through random region cropping. AWT~\cite{zhu2024awt} formulates the matching between randomly cropped visual regions and descriptions generated by LLMs as an optimal transport problem, significantly enhancing zero-shot recognition. In comparison, test-time prompt tuning~\cite{shu2022test,feng2023diverse,abdul2024align,imam2025test,sui2024just} fine-tunes image or textual prompts using learnable parameters. These parameters are trained by computing the matching entropy between randomly cropped visual regions and textual prompts, thus improving the alignment between the image and text. This study argues that filtering out irrelevant noise regions using the inherent knowledge of images aids in improving alignment between images and texts.

\section{Proposed Method}
\label{sec:method}

%-------------------------------------------------------------------------
 As illustrated in Figure \ref{fig:main}, our CPE method consists of two key components: Topology-Guided Synonymous Semantic Generation (TGSSG) and Category-Agnostic Discriminative Region Selection (CADRS). For each candidate class names, synonyms and descriptions are generated using large language models (LLMs) and input into the TGSSG to construct a comprehensive textual set. Given a test image, the activation map is extracted by the pre-trained vision model and fed into the CADRS to build a compact visual set. The set-to-set visual-textual matching is then applied to measure the predicted probability of the visual set corresponding to each class's textual set. A detailed description of these three modules is provided below.

%-------------------------------------------------------------------------
\subsection{Topology-Guided Synonymous Semantic Generation}
Textually, zero-shot generalization in VLMs generally is affected by the issue on lexical variation of textual prompts. Recent works~\cite{mirza2024meta,zhu2024awt} propose to use two-stage generation for textual descriptions via LLMs, which allow more nuanced semantic information and lead to competitive performance. However, existing methods focus solely on the given labels, while we argue that synonyms with related semantic contribute to a more comprehensive semantic representation of a concept (Figure \ref{fig:laion}). To address this issue, we propose Topology-Guided Synonymous Semantic Generation (TGSSG), which utilizes LLMs to generate synonyms and descriptions for category names. To mitigate noise caused by hallucinations in the LLM outputs, we filter irrelevant candidates using semantic ambiguity entropy and persistent homology analysis, thereby constructing comprehensive textual prompts sets.

For one of the candidate class names $C_k$, we first prompt an off-the-shelf LLM (e.g., Claude~\cite{anthropic2024claude}) as: \texttt{"Tell me in five words or less what are some common ways of referring to \{class\_name\}?"}. It generates synonyms set $\mathbf{S_k}=\{s_0, s_1,...\}$ for the class name $C_k$. However, hallucinations in LLMs often lead to suboptimal results~\cite{zhang2023siren,huang2023survey}. To ensure semantic consistency with the target class, we process the generated synonym set by analyzing its topology and removing outlier synonyms. Specifically, we map all synonyms to a feature set $\mathbf{F_k}=\{f_0, f_1,...\}$ using the CLIP text encoder and compute the semantic ambiguity entropy:
\begin{equation}
H_k(s_i) = -\sum_{j \neq i} \operatorname{d}(f_i,f_j) \log \operatorname{d}(f_i,f_j),
\end{equation}
where $\operatorname{d}(\cdot,\cdot)$ denotes the cosine similarity distance and $H_k(s_i)$ reflects the degree of semantic deviation of $s_i$ within the set $\mathbf{S_k}$. Next, we construct the Vietoris-Rips complex~\cite{mischaikow2004computational,zhu2013persistent} of the synonym feature set $\mathbf{F_k}$ and compute its persistent homology persistence $P_k$, which characterizes the topological compactness of the set. Finally, we retain synonyms that satisfy the condition:
\begin{equation}
\mathbf{\hat{S}_k} = \left\{ s_i \mid H_k(s_i)P_k < \mathbb{E}[P],\ s_i \in \mathbf{S_k} \right\},
\end{equation}
where $\mathbb{E}[P]$ represents the mean of persistence $P_k$ across all classes. The combined entropy and topological persistence filter out synonyms that deviate significantly from the primary semantic distribution, optimizing the reasonableness of semantic expression. Subsequently, we employ a two-step dataset-aware prompting strategy to generate a tailored description set $\mathbf{D_k}$ for each class label $C_k$ following~\cite{mirza2024meta,zhu2024awt}. By combining this description set $\mathbf{D_k}$ with the synonyms set $\mathbf{\hat{S}_k}$ through the template \texttt{"a photo of a \{synonym\}, \{description\}"} and encoding them with the CLIP text encoder, we obtain the comprehensive textual prompts set $\mathbf{T_k}$.

%-------------------------------------------------------------------------
\subsection{Category-Agnostic Discriminative Region Selection}
Visually, it is commonly assumed that diverse multi-scale image views are more suited for fine text alignment, compared to the whole image~\cite{zhu2024awt,li2024visual}. However, random image cropping and flipping strategy used in existing works~\cite{zhu2024awt,li2024visual} often introduce irrelevant noise. Although integration of cross-modal matching entropy can mitigate effect of these noise, it hardly fully eliminates irrelevant noise, thereby bringing the side effect on performance of text-image alignment (Figure \ref{fig:crop}). To address the issue of irrelevant visual noise introduced by random cropping, we propose Category-Agnostic Discriminative Region Selection (CADRS), which identifies discriminative regions through important activations and filters out background noise using statistical thresholds, thereby generating a compact visual prompts set.

Formally, given a test image $X \in \mathbb{R}^{3 \times H \times W}$, we utilize a pre-trained vision model (e.g., DINO~\cite{caron2021emerging}) to generate a category-agnostic attention map $A \in \mathbb{R}^{H \times W}$. Both the test image $X$ and the attention map $A$ undergo the same standard data augmentations, including random resized cropping and random flipping, producing augmented views $\{X_n\}_{n=1}^N$ and $\{A_n\}_{n=1}^N$, respectively. In practice, irrelevant noise views typically correspond to background regions, where activation values in the attention map are expected to be low. We hypothesize that these regions can be identified through the statistical properties of the attention map's activations. Specifically, we model the average activation values across all views $\{A_n\}_{n=1}^N$ as a Gaussian distribution, $N(\mu,\sigma^2)$, where $\mu$ and $\sigma$ represent the mean and standard deviation of the activation values, respectively. This model allows for the identification of views with low activations, which are often associated with irrelevant background noise. To quantitatively determine which views to discard, we apply a significance threshold using a two-standard deviation rule~\cite{bland1996statistics}. Specifically, views with activation values below $\mu - 2\sigma$ are considered outliers and discarded, while those above this threshold are retained. The remaining indices $\mathbf{I} = \left\{ i \mid \mathbb{E}[A_i] > \mu - 2\sigma\right\}$, where $\mathbb{E}[A_i]$ denotes the average activation of $A_i$, resulting in a set of selected views that are considered to contain meaningful information for text-image alignment. Together with the original image $X$, the compact visual prompts set is constructed as follows:
\begin{equation}
\mathbf{V} = \left\{ \phi(x) \mid x \in \{X\} \cup \{X_i\}_{i \in \mathbf{I}} \right\},
\end{equation}
where $\phi$ donates the CLIP image encoder. This method ensures that only the most relevant image regions, based on statistical significance, are retained for downstream tasks, thus enhancing the robustness and efficiency of text-image alignment.

%-------------------------------------------------------------------------
\subsection{Set-to-set Visual-textual Matching}
To achieve precise matching between our constructed comprehensive textual prompts and compact visual prompts for zero-shot prediction, we introduce two set-to-set matching strategies based on test-time adaptation (TTA) and optimal transport (OT), resulting in CPE-TTA and CPE-OT methods. Note that the pipelines of CPE-TTA and CPE-OT methods are illustrated in Figure \ref{fig:matching} in the Appendix.

\paragraph{CPE-TTA.} Test-time adaptation (TTA) enables dynamic adaptation of textual prompts to better align with visual content during inference. In our approach, TTA is used to dynamic adapt the matching between the visual set and textual set constructed in the previous steps. Specifically, for a test image, we replace random cropping in~\cite{sui2024just} with the compact multi-scale visual set $\mathbf{V}$. For each category's comprehensive textual set $\mathbf{T_k}$, we apply a learnable vector $l_k$ to uniformly shift the embeddings in the set at the channel level, as described in~\cite{sui2024just}. Next, we calculate predicted probability of each view in the image set $v_i \in \mathbf{V}$ with respect to each textual set $\mathbf{T_k}$ as 
\begin{equation}
p(k \mid v_i) = \frac{\exp\bigl(({\mathbb{E}[\mathbf{T_k}]}+l_k)^\top v_i / \tau\bigr)}{\sum_{j} \exp\bigl(({\mathbb{E}[\mathbf{T_j}]}+l_j)^\top v_i / \tau\bigr)}
\end{equation}
where $\mathbb{E}[\mathbf{T_k}]$ represents the average embedding of the textual set $\mathbf{T_k}$ and $\tau$ denotes the temperature scalar. Similar to \cite{sui2024just}, we select the $m$ distributions with lowest entropy of the batch and take the average. By minimizing the entropy of this marginal distribution, we update $l_k$ via a single-step gradient descent. During inference, the similarity between the center of the image set and the center of the shifted textual set is measured to perform zero-shot prediction.

\paragraph{CPE-OT.} Optimal Transport (OT), originating from the Monge problem~\cite{monge1781memoire}, provides a framework to measure similarity between distributions by finding the most efficient way to transform one distribution into another. In our context, we treat the visual set $\mathbf{V}$ and textual set $\mathbf{T_k}$ as two discrete distributions. Mathematically, we model each element in both sets as a mass located at its embedding position:
\begin{equation}
\alpha = \sum_{i=1}^{N} \mathbf{a}_i \delta_{v_i} \quad \text{and} \quad \beta^k = \sum_{j=1}^{M} \mathbf{b}_j \delta_{t_j^k},
\end{equation}
where $N$ and $M$ are number of samples in $\mathbf{V}$ and $\mathbf{T_k}$, $\delta$ denotes the Dirac function. $\mathbf{a}_i$ and $\mathbf{b}_j$ are the entropy-based importance weight of each element introduced in~\cite{zhu2024awt}. The transportation cost between any element $v_i$ from visual set $\mathbf{V}$ and any element $t_j^k$ from textual set $\mathbf{T_k}$ is quantified using the cosine distance between their embeddings. Therefore, we can adopt the Kantorovich relaxation~\cite{kantorovich2006translocation} to form the optimal transport problem between the two sets as:
\begin{equation}\label{eq:ot}
\mathbf{P}^*_{k} = \underset{\mathbf{P} \in \mathbb{R}^{M \times N}}{\arg \min} \sum_{i=1}^{M} \sum_{j=1}^{N} \mathbf{C}_{ij} \mathbf{P}_{ij}
\qquad
\text{s.t.} \quad \mathbf{P}\mathbf{1}_{M} = \mathbf{a} \quad \text{and} \quad \mathbf{P}^{\mathrm{T}}\mathbf{1}_{N} = \mathbf{b}.
\end{equation}
Here $\mathbf{C} \in \mathbb{R}^{M \times N}$ defines the cost matrix. Eq. \ref{eq:ot} can be efficiently approximated using the Sinkhorn Algorithm~\cite{cuturi2013sinkhorn}. Consequently, the classification probability can be expressed as:
\begin{equation}
p(k \mid \mathbf{V}) = \frac{\exp( \mathbf{P}^*_{k} / \tau)}{\sum_{j} \exp(\mathbf{P}^*_{j} / \tau)},
\end{equation}
where $\mathbf{P}^*_{k}$ represents the optimal transport cost from $\mathbf{V}$ to $\mathbf{T_k}$ and $\tau$ denotes the temperature scalar.

\section{Experiments}
\label{sec:experimental}

%-------------------------------------------------------------------------
\subsection{Implementation Details}
In this section, we utilize CLIP ViT-B/16 as the basic model for comparing all methods. To generate synonyms, we leverage the public web API of Claude-3.5-Sonnet~\cite{anthropic2024claude} by setting the number of synonyms $M^*$ to 5. Descriptions are generated using a script based on the code repository provided in \cite{mirza2024meta,zhu2024awt} by using the identical configurations. The number of candidate augmented images $N^*$ is initialized by 100. Activation maps are generated using Attention Rollout~\cite{clark2019does} on DINO~\cite{caron2021emerging}. A detailed discussion of the hyperparameters is provided in the ablation study. All experiments are conducted in PyTorch and run on a PC with dual NVIDIA RTX 3090 GPUs.

\begin{table*}[t]
    \centering
    \footnotesize
    \setlength{\tabcolsep}{4pt}  % Adjusts column padding
    \renewcommand{\arraystretch}{1}  % Adjust row height
    \begin{tabular}{l|*{10}{p{0.71cm}}|>{\centering\arraybackslash}p{1.2cm}}
    \toprule
    & \rotatebox{60}{\textbf{Flowers}} & \rotatebox{60}{\textbf{DTD}} & \rotatebox{60}{\textbf{Pets}} & \rotatebox{60}{\textbf{Cars}} & \rotatebox{60}{\textbf{UCF}} & \rotatebox{60}{\textbf{CalTech}} & \rotatebox{60}{\textbf{Food}} & \rotatebox{60}{\textbf{SUN}} & \rotatebox{60}{\textbf{Aircraft}} & \rotatebox{60}{\textbf{SAT}} & \rotatebox{60}{\textit{\textbf{Avg.}}} \\
    \midrule
    \midrule
    CLIP~\cite{radford2021learning} & 67.44 & 44.27 & 88.25 & 65.48 & 65.13 & 93.35 & 83.65 & 62.59 & 23.67 & 42.01 & 63.58 \\
    \midrule
    CoOp~\cite{zhou2022learning} & 68.71 & 41.92 & 89.14 & 64.51 & 66.55 & 93.70 & 85.30 & 64.15 & 18.47 & 46.39 & 63.88 \\
    CoCoOp~\cite{zhou2022conditional} & 71.88 & 45.73 & 90.14 & 65.32 & 68.21 & 94.43 & 86.06 & 67.36 & 22.94 & 45.37 & 65.74 \\
    MaPLe~\cite{khattak2023maple} & 72.23 & 46.49 & 90.49 & 65.57 & 68.69 & 93.53 & 86.20 & 67.01 & 24.74 & 48.06 & 66.30 \\
    PLOT++~\cite{chen2022plot} & 69.10 & 38.42 & 90.49 & 61.20 & 68.94 & 91.32 & 86.07 & 61.59 & 24.84 & 49.90 & 64.19 \\
    POMP~\cite{ren2023prompt} & 72.72 & 44.44 & 89.05 & 66.70 & 68.44 & 94.65 & 86.28 & 67.27 & 25.47 & 52.65 & 66.77 \\
    ProVP-Ref~\cite{xu2025progressive} & 71.62 & 45.97 & 91.58 & 65.29 & 67.72 & 93.79 & 86.17 & 66.29 & 24.51 & 51.95 & 66.49 \\
    \midrule
    TPT~\cite{shu2022test} & 68.98 & 47.75 & 87.79 & 66.87 & 68.04 & 94.16 & 84.67 & 65.50 & 24.78 & 42.44 & 65.10 \\
    DiffTPT~\cite{feng2023diverse} & 70.10 & 47.00 & 88.22 & 67.01 & 68.22 & 92.49 & \textbf{87.23} & 65.74 & 25.60 & 43.13 & 65.47 \\
    TTL~\cite{imam2025test} & 70.48 & 46.69 & 88.72 & 67.96 & 69.20 & 93.63 & 85.05 & 66.32 & 23.82 & 42.02 & 65.39 \\
    PromptAlign~\cite{abdul2024align}& 72.39 & 47.24 & 90.76 & 68.50 & 69.47 & 94.01 & 86.65 & 67.54 & 24.80 & 47.86 & 66.92 \\
    TPS~\cite{sui2024just} & 71.54 & 50.47 & 87.35 & 69.06 & 71.00 & 95.09 & 85.23 & 68.98 & 26.34 & 44.48 & \textcolor{blue}{66.95} \\
    \midrule
    % DCLIP~\cite{menon2022visual} & 70.52 & 49.82 & 87.30 & 66.70 & 70.34 & 93.96 & 84.50 & 67.47 & 24.81 & 44.37 & 65.98 \\
    CuPL~\cite{pratt2023does} & 71.30 & 44.56 & 89.13 & 65.29 & 66.83 & 92.98 & 86.11 & 62.59 & 24.90 & 47.84 & 65.15 \\
    DCLIP~\cite{menon2022visual} & 70.85 & 44.98 & 88.85 & 64.08 & 67.12 & 94.60 & 85.05 & 67.99 & 24.30 & 54.84 & 66.27 \\
    WaffleCLIP~\cite{roth2023waffling} & 72.35 & 45.21 & 89.95 & 63.57 & 67.19 & 94.02 & 86.68 & 67.23 & 25.39 & 55.07 & 66.67 \\
    SuS-X~\cite{udandarao2023sus} & 73.81 & 54.55 & 90.57 & 66.13 & 66.59 & 93.96 & 86.08 & 67.73 & 28.68 & 57.49 & 68.56 \\
    REAL~\cite{parashar2024neglected} & 73.20 & 51.12 & 91.41 & 66.45 & 65.40 & 90.22 & 83.71 & 62.61 & 24.69 & 54.44 & 66.33 \\
    MPVR~\cite{mirza2024meta} & 76.90 & 56.10 & 89.90 & 65.40 & 70.90 & 94.10 & 86.40 & 68.80 & 28.00 & 59.60 & 69.61 \\
    AWT~\cite{zhu2024awt} & 75.07 & 55.56 & 92.53 & 69.93 & 72.51 & 95.54 & 85.54 & 70.58 & 29.22 & 58.61 & \textcolor{blue}{70.51} \\
    \midrule
    \rowcolor{gray!20} CPE-TTA & 77.83 & 55.79 & 89.29 & 66.96 & 72.51 & 95.25 & 85.86 & 70.21 & 29.70 & 54.81 & 69.82\textcolor{blue}{\scriptsize $\uparrow$2.87} \\
    \rowcolor{gray!20} CPE-OT & \textbf{84.25} & \textbf{57.80} & \textbf{92.59} & \textbf{70.17} & \textbf{73.06} & \textbf{95.78} & 86.06 & \textbf{70.65} & \textbf{32.67} & \textbf{62.59} & \textbf{72.56}\textcolor{blue}{\scriptsize $\uparrow$2.05} \\
    \bottomrule
    \end{tabular}
    \caption{\textbf{Zero-shot image classification.} We report top-1 accuracy (\%) for each dataset. The best accuracies are highlighted in \textbf{bold}, and the average for comparison with CPE is presented in \textcolor{blue}{blue}.}
    \label{tab:image}
\end{table*}

%-------------------------------------------------------------------------
\subsection{Zero-Shot Image Classification}
We conduct zero-shot image classification on 10 widely used datasets: Oxford Flowers~\cite{nilsback2008automated}, Oxford Pets~\cite{parkhi2012cats}, Stanford Cars~\cite{krause20133d}, Food101~\cite{bossard2014food}, and FGVC Aircraft~\cite{maji2013fine} for fine-grained classification, DTD~\cite{cimpoi2014describing} for texture classification, UCF101~\cite{soomro2012ucf101} for action recognition, Caltech101~\cite{fei2004learning} for generic object recognition, SUN397~\cite{xiao2010sun} for scene recognition, and EuroSAT~\cite{helber2019eurosat} for satellite recognition. For a thorough evaluation, we primarily compare four distinct categories of zero-shot methods:
% We conduct zero-shot image classification on 10 widely used datasets: Oxford Flowers~\cite{nilsback2008automated}, DTD~\cite{cimpoi2014describing}, Oxford Pets~\cite{parkhi2012cats}, Stanford Cars~\cite{krause20133d}, UCF101~\cite{soomro2012ucf101}, Caltech101~\cite{fei2004learning}, Food101~\cite{bossard2014food}, SUN397~\cite{xiao2010sun}, Aircraft~\cite{maji2013fine}, and EuroSAT~\cite{helber2019eurosat}. For a thorough evaluation, we primarily compare four distinct categories of zero-shot methods:
(1) Baseline: The standard CLIP model~\cite{radford2021learning} based on the ViT-B/16 architecture, using a conventional prompt template.
(2) Prompt learning: Methods those incorporate additional data for model fine-tuning, such as CoOp~\cite{zhou2022learning}, CoCoOp~\cite{zhou2022conditional}, MaPLe~\cite{khattak2023maple}, PLOT++~\cite{chen2022plot}, POMP~\cite{ren2023prompt}, and ProVP-Ref~\cite{xu2025progressive}.
(3) Test-time prompt tuning: Methods those optimize visual or textual prompts during inference, including TPT~\cite{shu2022test}, DiffTPT~\cite{feng2023diverse}, TTL~\cite{imam2025test}, PromptAlign~\cite{abdul2024align}, and TPS~\cite{sui2024just}.
(4) Prompt enhancement: Methods those leverage LLMs to generate enriched prompts, such as CuPL~\cite{pratt2023does}, DCLIP~\cite{menon2022visual}, WaffleCLIP~\cite{roth2023waffling}, SuS-X~\cite{udandarao2023sus}, REAL~\cite{parashar2024neglected}, MPVR~\cite{mirza2024meta}, and AWT~\cite{zhu2024awt}.

The compared results are given in Table \ref{tab:image}, from it we can draw the following conclusions. 
(1) Compared to prompt learning methods that perform well in few-shot scenarios, our method achieves a more than 6\% improvement in average accuracy across 10 datasets, even though prompt learning methods fine-tune with additional data from the ImageNet dataset. This demonstrates that, unlike few-shot scenarios in VLMs, additional data fine-tuning is not necessary for zero-shot generalization.
(2) In comparison to test-time prompt tuning methods, our CPE-TTA achieves an average accuracy 2.87\% higher than the previous best result, TPS. This indicates that enhancing both visual and textual prompts is equally effective for test-time prompt tuning.
(3) Compared to prompt enhancement methods, our CPE-OT further enhances the prompts, outperforming the previous best, AWT, by 2.05\% in average accuracy. This suggests that comprehensive textual prompts and compact visual prompts can significantly improve zero-shot recognition.
(4) Our method outperforms all existing approaches with a significant advantage, achieving state-of-the-art performance in 9 out of 10 datasets, demonstrating the superiority of the proposed method.

\begin{table}[t]
    \footnotesize
    \renewcommand{\arraystretch}{1}
    \begin{minipage}[t]{0.6\textwidth}
        \centering
        \setlength{\tabcolsep}{2pt}
        \begin{tabular}{>{\raggedright\arraybackslash}p{2cm}|>{\centering\arraybackslash}p{0.9cm}|*{4}{>{\centering\arraybackslash}p{0.9cm}}|>{\centering\arraybackslash}p{0.9cm}}
        \toprule
        % & \rotatebox{60}{\textbf{IN-1k}} & \rotatebox{60}{\textbf{IN-A}} & \rotatebox{60}{\textbf{IN-V2}} & \rotatebox{60}{\textbf{IN-R}} & \rotatebox{60}{\textbf{IN-K}} & \rotatebox{60}{\textit{\textbf{OOD}}} \\
        & \textbf{IN-1k} & \textbf{IN-A} & \textbf{IN-V2} & \textbf{IN-R} & \textbf{IN-K} & \textit{\textbf{OOD}} \\
        \midrule
        CLIP~\cite{radford2021learning} & 66.74 & 47.74 & 60.75 & 73.98 & 46.13 & 57.15 \\
        TPT~\cite{shu2022test} & 68.98 & 54.77 & 63.45 & 77.06 & 47.94 & 60.81 \\
        DiffTPT~\cite{feng2023diverse} & 70.30 & 55.68 & 65.10 & 75.00 & 46.80 & 60.65 \\
        TPS~\cite{sui2024just} & 71.45 & \textbf{60.61} & 64.91 & 80.20 & 50.88 & 64.15 \\
        CuPL~\cite{pratt2023does} & 69.62 & 50.72 & 63.27 & 77.05 & 49.02 & 60.02 \\
        DCLIP~\cite{menon2022visual} & 68.55 & 49.07 & 61.80 & 75.13 & 47.97 & 58.49 \\
        WaffleCLIP~\cite{roth2023waffling} & 68.81 & 50.78 & 62.54 & 77.49 & 49.10 & 59.98 \\
        REAL~\cite{parashar2024neglected} & 68.50 & 50.04 & 61.97 & 77.69 & 48.19 & 59.47 \\
        MPVR~\cite{mirza2024meta} & 69.70 & - & 63.40 & 78.20 & 50.60 & - \\
        AWT~\cite{zhu2024awt} & 71.32 & 60.33 & 65.15 & 80.64 & 51.60 & 64.43 \\
        \midrule
        \rowcolor{gray!20} CPE-OT & \textbf{71.74} & 60.12 & \textbf{65.48} & \textbf{80.69} & \textbf{52.76} & \textbf{64.76} \\
        \bottomrule
        \end{tabular}
        \vspace*{6pt}
        \caption{\textbf{Zero-shot out-of-distribution generalization.}}
        \label{tab:ood}
    \end{minipage}
    \hspace{0.02\textwidth}
    \begin{minipage}[t]{0.32\textwidth}
        \centering
        \setlength{\tabcolsep}{1pt}
        % \begin{tabular}{>{\raggedright\arraybackslash}p{1.8cm}|*{3}{>{\centering\arraybackslash}p{0.8cm}}}
        \begin{tabular}{>{\raggedright\arraybackslash}p{2.3cm}|>{\centering\arraybackslash}p{0.85cm}|>{\centering\arraybackslash}p{0.85cm}|>{\centering\arraybackslash}p{0.85cm}}
        \toprule
        & \textbf{UCF} & \textbf{HM} & \textbf{K600} \\
        \midrule
        ActionCLIP~\cite{wang2021actionclip} & 77.5 & 48.2 & 62.5 \\
        X-CLIP~\cite{ni2022expanding} & 72.0 & 44.6 & 65.2 \\
        AIM~\cite{yang2022aim} & 79.4 & 50.3 & 66.7 \\
        ST-Adapter~\cite{pan2022st} & 77.6 & 51.1 & 60.2 \\
        Vita-CLIP~\cite{wasim2023vita} & 75.0 & 48.6 & 67.4 \\
        ViFi-CLIP~\cite{rasheed2023fine} & 76.8 & 51.3 & 71.2 \\
        AdaptFormer~\cite{chen2022adaptformer} & 80.3 & 51.0 & 67.0 \\
        Open-VCLIP~\cite{weng2023open} & 83.4 & 53.9 & 73.0 \\
        FROSTER~\cite{huang2024froster} & 84.8 & 54.8 & 74.8 \\
        AWT~\cite{zhu2024awt} & 85.2 & 57.2 & 76.1 \\
        \midrule
        \rowcolor{gray!20} CPE-OT & \textbf{85.7} & \textbf{57.9} & \textbf{77.3} \\
        \bottomrule
        \end{tabular}
        \vspace*{6pt}
        \caption{\textbf{ZS video recognition.}}
        \label{tab:video}
    \end{minipage}
\end{table}
\begin{table*}[t]
    \footnotesize
    \begin{minipage}[t]{0.39\textwidth}
        \centering
        \renewcommand{\arraystretch}{1}
        \setlength{\tabcolsep}{1pt}
        \subcaption{Effect of key components}
        \begin{tabular}{>{\raggedright\arraybackslash}p{2.2cm}|>{\centering\arraybackslash}p{1.2cm}>{\centering\arraybackslash}p{1.2cm}}
        \toprule
        Step & IN-1k & Avg.(10) \\
        \midrule
        AWT (baseline) & 71.28 & 70.44 \\
        +V & 71.35 & 70.49 \\
        +T w/o filter & 71.46 & 71.81 \\
        +T & 71.58 & 72.33 \\
        \rowcolor{gray!20} +V+T & \textbf{71.74} & \textbf{72.56} \\
        \bottomrule
        \end{tabular}
        \label{tab:component}
    \end{minipage}
    \hspace{0.01\textwidth}
    \begin{minipage}[t]{0.27\textwidth}
        \centering
        \renewcommand{\arraystretch}{0.85}
        \setlength{\tabcolsep}{1pt}
        \subcaption{Effect of number $M^*$}
        \begin{tabular}{>{\centering\arraybackslash}p{1cm}|>{\centering\arraybackslash}p{1.2cm}>{\centering\arraybackslash}p{1.2cm}}
        \toprule
        $M^*$ & IN-1k & Avg.(10) \\
        \midrule
        1 & 68.84 & 70.62 \\
        2 & 70.72 & 71.67 \\
        3 & 71.06 & 72.40 \\
        4 & 71.37 & 72.52 \\
        \rowcolor{gray!20} 5 & \textbf{71.74} & \textbf{72.56} \\
        6 & 71.68 & 72.53 \\
        \bottomrule
        \end{tabular}
        \label{tab:synonyms}
    \end{minipage}
    \hspace{0.01\textwidth}
    \begin{minipage}[t]{0.3\textwidth}
        \centering
        \renewcommand{\arraystretch}{1}
        \setlength{\tabcolsep}{1pt}
        \subcaption{Effect of number $N^*$}
        \begin{tabular}{>{\centering\arraybackslash}p{1.2cm}|>{\centering\arraybackslash}p{1.2cm}>{\centering\arraybackslash}p{1.2cm}}
        \toprule
        $N^*$ & IN-1k & Avg.(10) \\
        \midrule
        25 & 71.55 & 72.39 \\
        50 & 71.43 & 72.30 \\
        \rowcolor{gray!20} 100 & \textbf{71.74} & \textbf{72.56} \\
        150 & 71.73 & 72.43 \\
        200 & 71.69 & 72.41 \\
        \bottomrule
        \end{tabular}
        \label{tab:views}
    \end{minipage}
    \caption{\textbf{Ablation study} across ImageNet and 10 image classification datasets. The default configuration is colored \colorbox{gray!20}{gray}. T and V indicates textual and visual prompt enhancement, respectively.}
\end{table*}

%-------------------------------------------------------------------------
\subsection{Zero-Shot Out-of-Distribution Generalization}
We assess the domain generalization ability of our CPE-OT method on 5 natural distribution datasets including ImageNet~\cite{deng2009imagenet} and its out-of-distribution variants ImageNet-A~\cite{hendrycks2021natural}, ImageNet-V2~\cite{recht2019imagenet}, ImageNet-R~\cite{hendrycks2021many}, and ImageNet-Sketch~\cite{wang2019learning}. Table \ref{tab:ood} compares the results of various methods on ImageNet and four OOD benchmarks, where we can obverse that (1) our CPE-OT method outperforms all existing approaches on ImageNet, achieving an improvement of 0.42\% over the recently proposed state-of-the-art (AWT). 
(2) Besides, our CPE-OT achieves the best performance on three out of four OOD datasets, and obtains the highest average score, surpassing the previous best results by 0.33\%. These improvement over AWT owes to our constrained visual and textual prompt enhancement helps to handle complex OOD data. (3) Note that our CPE-OT  achieves a substantial improvement of 1.16\% on ImageNet‑Sketch, owing to the uniform backgrounds and well‑defined object contours in hand‑drawn sketches, which facilitate the identification and localization of discriminative regions. In contrast, performance of CPE-OT  on ImageNet‑A is modest,  which may be caused by that natural adversarial examples introduce distracting artifacts that hinder the model’s discrimination. These results above clearly demonstrate that our CPE-OT is well generalized to OOD settings.

%-------------------------------------------------------------------------
\subsection{Zero-Shot Video Recognition}
Following the setup of AWT~\cite{zhu2024awt}, we evaluate the zero-shot generalization capability of our method for video action recognition using three widely used datasets: UCF101~\cite{soomro2012ucf101}, HMDB51~\cite{kuehne2011hmdb}, and Kinetics-600~\cite{carreira2018short}. For each dataset, we assess the model performance using three official or commonly used splits and report the average result across all splits. To capture temporal dynamics in visual features, we also follow Open-VCLIP~\cite{weng2023open} to use neighbor-frame attention and fine-tune CLIP on Kinetics-400. Textual prompt enhancement is the same for zero-shot image classification task.

In Table \ref{tab:video}, we compare our method with existing CLIP-based zero-shot video action recognition approaches~\cite{wang2021actionclip,ni2022expanding,yang2022aim,pan2022st,wasim2023vita,rasheed2023fine,chen2022adaptformer,weng2023open,huang2024froster}. According to the results, we have two observations. (1) Although our method is not specifically designed for video tasks, it achieves state-of-the-art performance, surpassing the previous best AWT by 0.5\%, 0.7\% and 1.2\% on UCF101, HMDB51 and Kinetics-600, respectively. These results demonstrate that our approach is effective for video understanding tasks.
(2) Compared to AWT, our method improves zero-shot video recognition solely by constructing comprehensive textual prompts set, demonstrating the significance of textual semantics on video understanding.

%-------------------------------------------------------------------------
\subsection{Ablation Study}
\paragraph{Effect of key components.} Table \ref{tab:component} presents an analysis for effect of key components on our method. Initially, we separately enhance the textual and visual components. The results (rows 2 and 4) indicate that enhancing the textual prompts yields more significant improvements than enhancing the visual prompts. Additionally, the results (rows 3 and 4) suggest that filtering the generated synonym sets is essential. Finally, we enhance both the textual and the visual prompts simultaneously, and the results (row 4) demonstrate that this combination produces the best performance.

\paragraph{Effect of hyperparameters.} Table \ref{tab:synonyms} and \ref{tab:views} present study on effect of hyperparameters in textual and visual prompts, including number of \textbf{generated synonyms} $M^*$ and number of \textbf{augmented image views} $N^*$. For the number of generated synonyms, performance improves along the increased number and stabilizes at $M^*=5$, indicating small number of synonyms is insufficient to capture the full textual semantic. For number of augmented image views, we found that augmentation of $N^*=100$ yields the best results. Particularly, too few augmentations fail to cover all useful characteristics, while too many augmentations will introduce excessive noise.

%-------------------------------------------------------------------------
\section{Conclusions}
In this paper, we proposed a constrained prompt enhancement method to improve zero-shot generalization of VLMs, whose core idea is to perform a set-to-set matching between comprehensive textual prompts and compact visual prompts for improving visual-textual alignment. To this end, a topology-guided synonymous semantic generation and a category-agnostic discriminative region selection are presented to construct comprehensive textual prompts and generated compact visual prompts, respectively. Subsequently, two set-to-set matching strategies are introduced to achieve effective visual-textual alignment. Extensive experiments demonstrate the effectiveness of our approach. We hope our analysis on completeness of textual prompts and compactness of visual prompts can encourage further research on improving generalization of VLMs.

%Bibliography
{\small
\bibliographystyle{unsrtnat}
\setlength{\bibsep}{4pt}
\bibliography{references}
}

%%%%%%%%%%%%%%%%%%%%%%%%%%%%%%%%%%%%%%%%%%%%%%%%%%%%%%%%%%%%
\newpage
\appendix

%-------------------------------------------------------------------------
\section{Generalization across Various Architectures}
\begin{figure}[h]
    \centering
    \begin{minipage}[b]{0.39\textwidth}
        \centering
        \includegraphics[width=\textwidth]{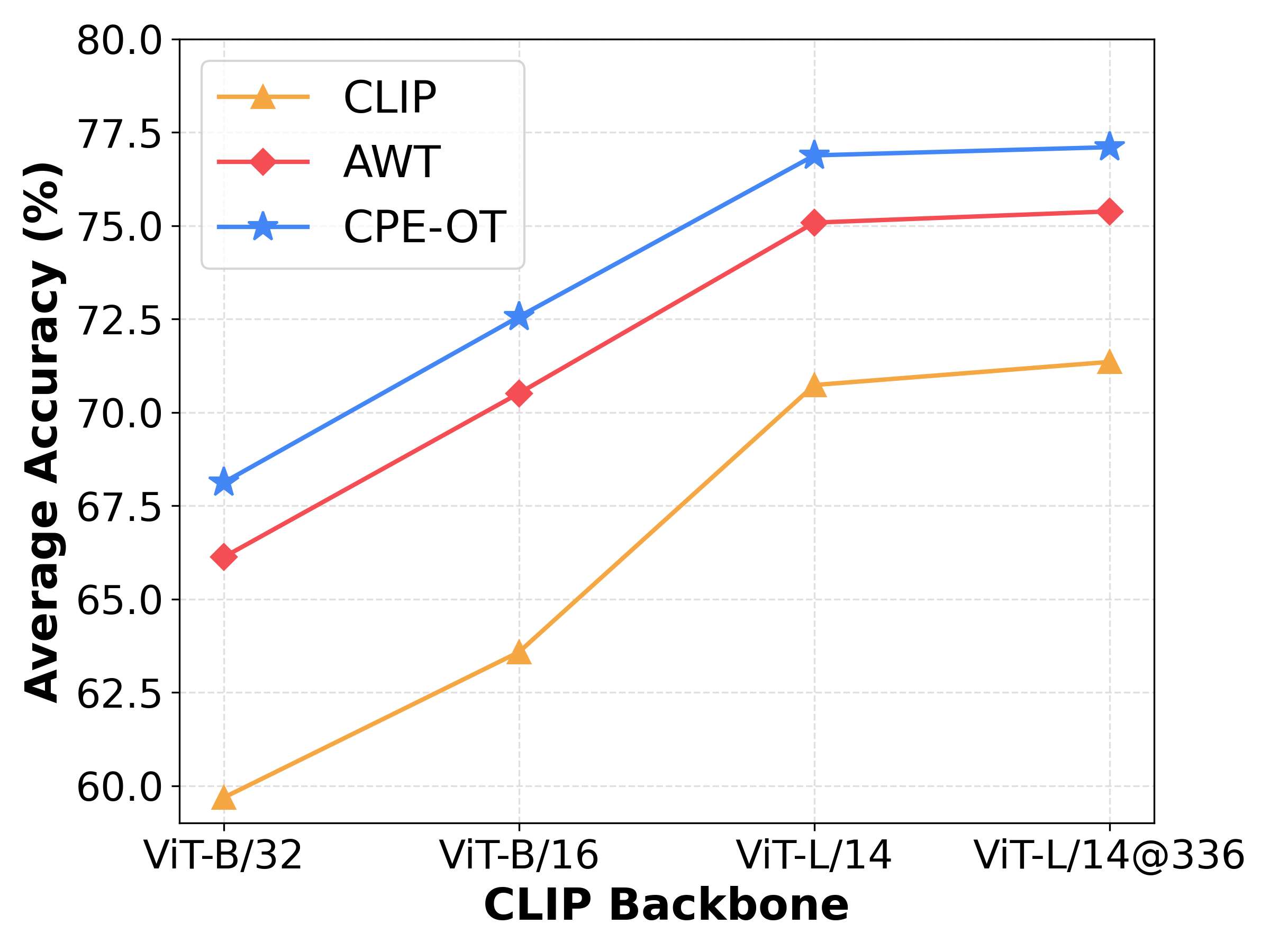}  
        \subcaption{Comparison across different backbones.}
        \label{fig:clip}
    \end{minipage}
    \hfill
    \begin{minipage}[b]{0.585\textwidth} 
        \centering
        \includegraphics[width=\textwidth]{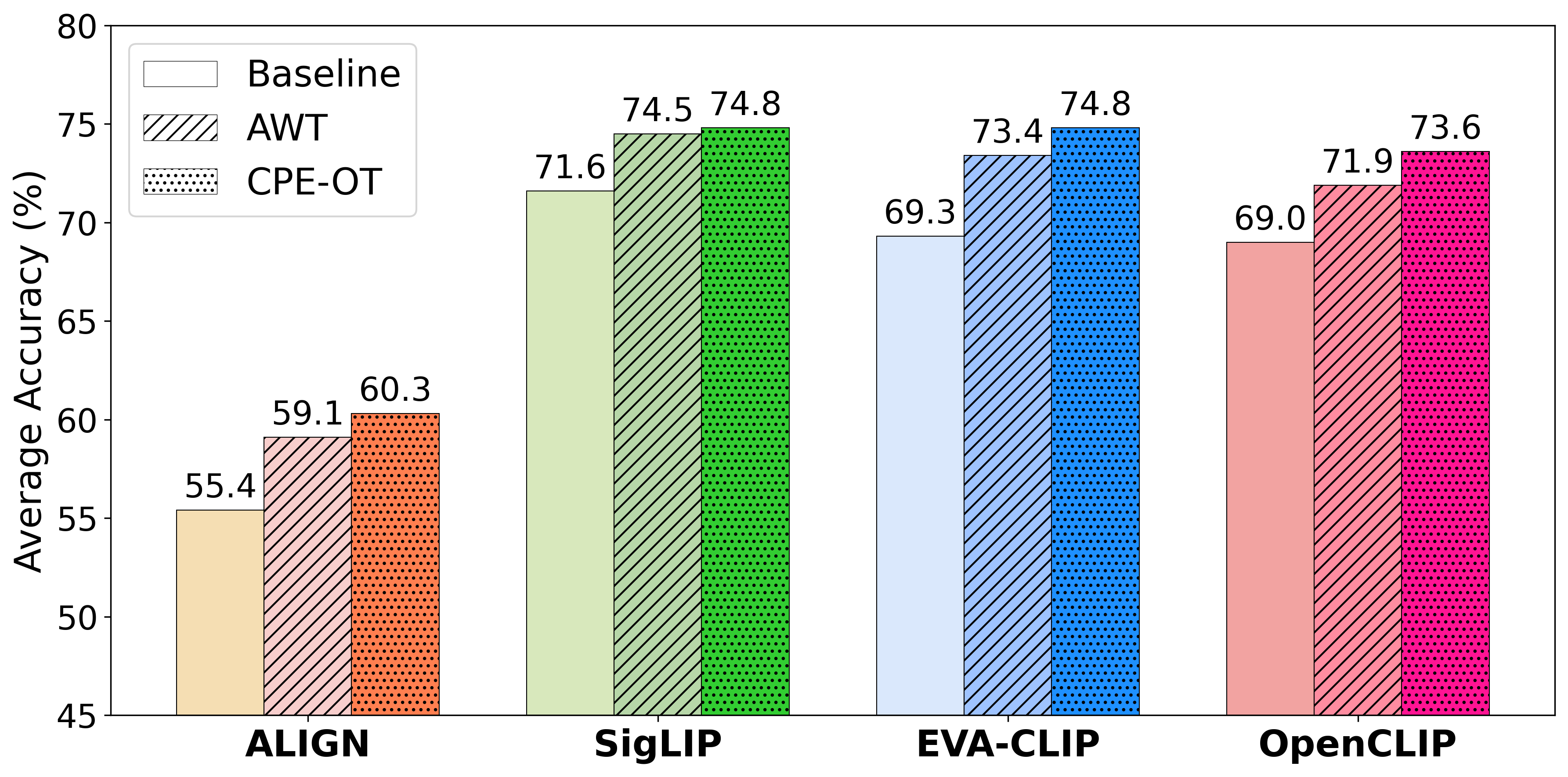}
        \subcaption{Generalization to different VLMs.}
        \label{fig:vlm}
    \end{minipage}
    \caption{\textbf{Generalizing across architectures}. Average top-1 accuracy (\%) on 10 zero-shot image datasets is reported.}
\end{figure}

We conduct evaluations across various CLIP architectures, exploring the model's scalability from ViT-B/32 to ViT-L/14@336. As shown in Figure \ref{fig:clip}, our CPE-OT consistently outperforms AWT by approximately 2\%. Additionally, we assess the model's generalization on four other VLMs: ALIGN~\cite{jia2021scaling}, SigLIP~\cite{zhai2023sigmoid}, EVA-CLIP~\cite{sun2023eva}, and OpenCLIP~\cite{cherti2023reproducible}, with results shown in Figure \ref{fig:vlm}. AWT achieves a 3\% improvement over the baseline, while our CPE-OT provides an additional 1\% boost over AWT. These results demonstrate that our method consistently achieves performance improvements across all evaluated scenarios, highlighting its generalization.

\section{Efficiency Analysis}
In Table \ref{tab:efficiency}, we compare the token consumption for text generation and the inference time during testing. Compared to the current competitive two-step dataset-aware prompting strategy (rows 5 and 6), our method requires only an additional 1\% of tokens to generate synonyms, while achieving a more than 2\% improvement in zero-shot recognition performance. At the same time, our method increases inference time by only 0.5\% compared to AWT. These results demonstrate that our approach significantly enhances zero-shot task performance while maintaining high inference efficiency and computational resource utilization.

\begin{table*}[h]
    \centering
    \footnotesize
    \renewcommand{\arraystretch}{1}
    \begin{tabular}{>{\raggedright\arraybackslash}p{1.5cm}>{\centering\arraybackslash}p{2cm}*{2}{>{\centering\arraybackslash}p{2cm}}>{\centering\arraybackslash}p{2cm}>{\centering\arraybackslash}p{2cm}}
        \toprule
        \multirow{3}{*}{Model}
        & \multicolumn{2}{c}{LLMs Generator Requirements}
        & \multirow{3}{*}{Token}
        & \multirow{3}{*}{Inference Time}
        & \multirow{3}{*}{Accuracy} \\
        \cmidrule(lr){2-3}
        & Words & Sentences & \multirow{2}{*}{Usage} & \multirow{2}{*}{(s)} & \multirow{2}{*}{(\%)} \\
        & (10K) & (500K) & & & \\
        \midrule
        CLIP~\cite{radford2021learning}    & - & - & -     & 0.035 & 63.58 \\
        CuPL~\cite{pratt2023does}          & - & 1 & 500K  & 0.035 & 65.15 \\
        REAL~\cite{parashar2024neglected}  & 1 & - & 10K   & 0.035 & 66.33 \\
        TPS~\cite{sui2024just}             & - & - & -     & 0.186 & 66.95 \\
        MPVR~\cite{mirza2024meta}          & - & 2 & 1000K & 0.035 & 69.61 \\
        AWT~\cite{zhu2024awt}              & - & 2 & 1000K & 0.103 & 70.51 \\
        \rowcolor{gray!20} CPE-OT          & 1 & 2 & 1010K & 0.108 & \textbf{72.56} \\
        \bottomrule
    \end{tabular}
    \caption{Efficiency analysis across different methods. For every 1K categories, LLMs generate approximately 10K tokens for words and 500K tokens for sentences (refer to \cite{parashar2024neglected}). Inference Time reports the average inference time per image (in seconds).}
    \label{tab:efficiency}
\end{table*}

%-------------------------------------------------------------------------
\newpage
\section{Pipeline of Set-to-set Visual-textual Matching}
\begin{figure}[h]
    \centering
    \begin{minipage}[b]{0.44\textwidth}
        \centering
        \includegraphics[width=\textwidth]{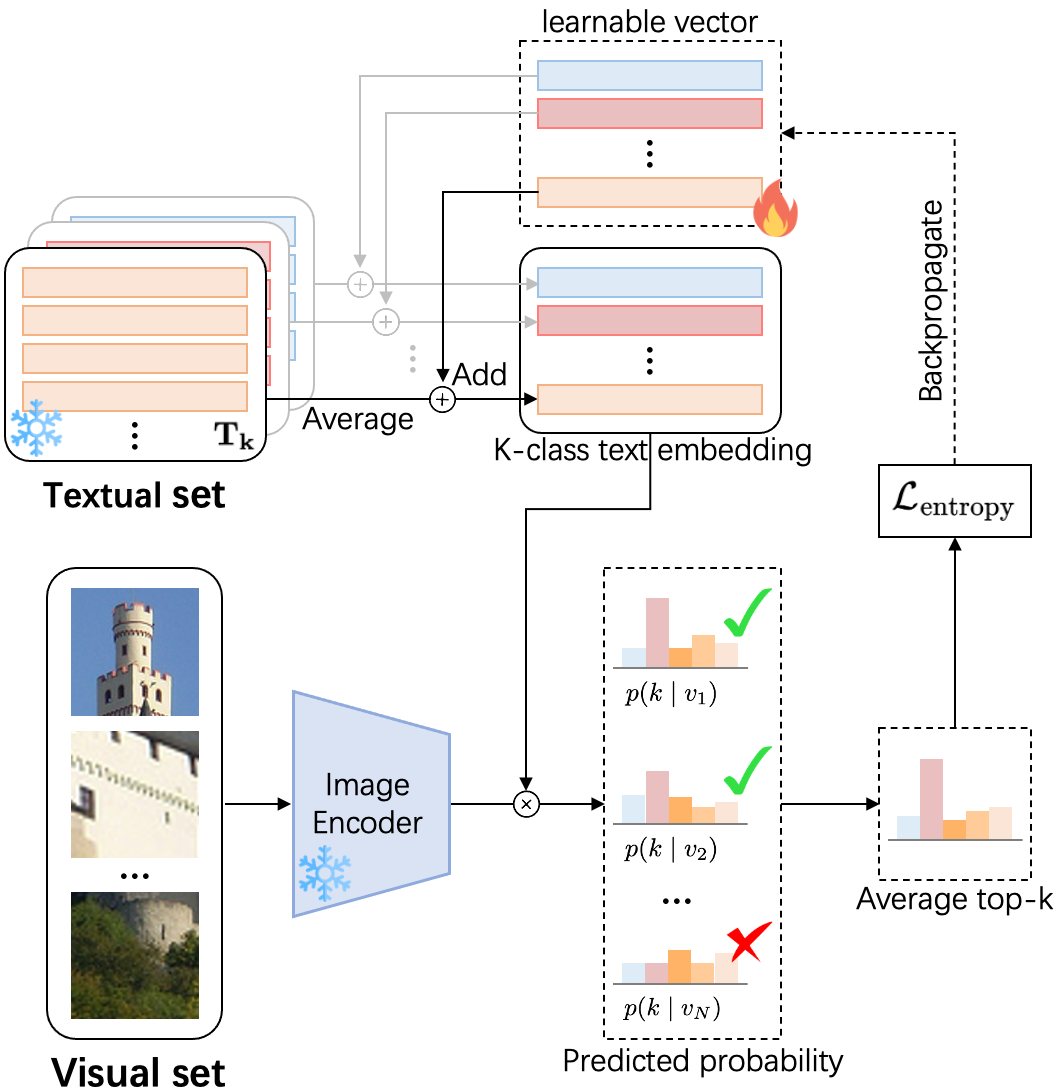}  
        \subcaption{CPE-TTA}
    \end{minipage}
    \hfill
    \begin{minipage}[b]{0.46\textwidth} 
        \centering
        \includegraphics[width=\textwidth]{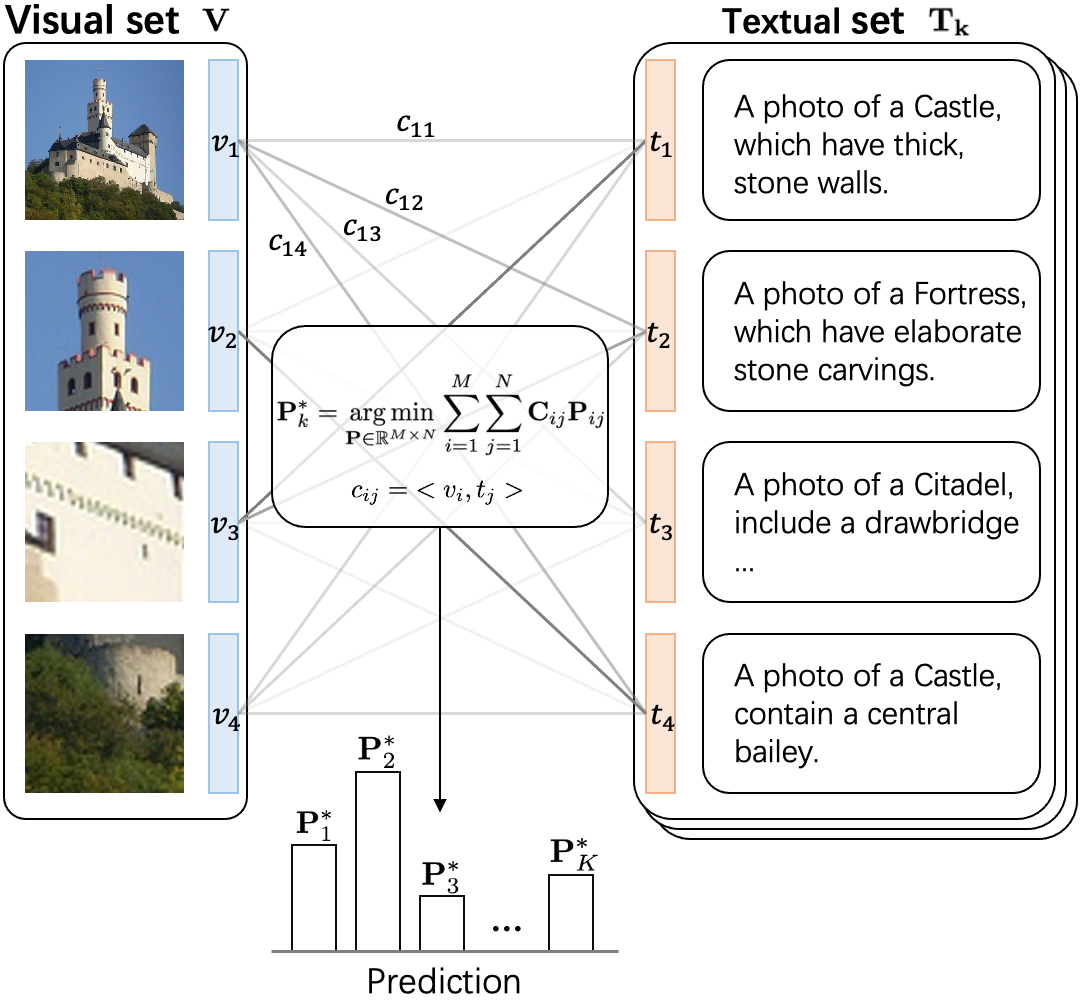}
        \subcaption{CPE-OT}
    \end{minipage}
    \caption{\textbf{Pipeline of CPE-TTA and CPE-OT.}}
    \label{fig:matching}
\end{figure}

%-------------------------------------------------------------------------
\section{Additional Details}

\subsection{Error bar analysis}
\label{sec:bar}
We conducted an analysis of error bars across 15 image datasets, each evaluated with three different random seeds to ensure statistical robustness. The results, presented in Table \ref{tab:bar}, show that our method consistently achieves robust performance with minimal variability on most datasets.

\begin{table*}[h]
    \footnotesize
    \renewcommand{\arraystretch}{1}
    \begin{minipage}{\textwidth}
        \centering
        \setlength{\tabcolsep}{2pt}
        \begin{tabular}{>{\centering\arraybackslash}p{1.8cm}|*{5}{>{\centering\arraybackslash}p{2.2cm}}}
        \toprule
        \textbf{Method} & \textbf{Oxford Flowers} & \textbf{DTD} & \textbf{Oxford Pets} & \textbf{Stanford Cars} & \textbf{UCF101} \\
        \midrule
        CLIP & 67.44 & 44.27 & 88.25 & 65.48 & 65.13 \\
        \rowcolor{gray!20} CPE-OT & 84.14$\pm$0.13 & 57.84$\pm$0.07 & 92.44$\pm$0.13 & 69.99$\pm$0.16 & 73.05$\pm$0.01 \\
        \bottomrule
        \end{tabular}
    \end{minipage}
    \vfill
    \begin{minipage}{\textwidth}
        \centering
        \setlength{\tabcolsep}{2pt}
        \begin{tabular}{>{\centering\arraybackslash}p{1.8cm}|*{5}{>{\centering\arraybackslash}p{2.2cm}}}
        \toprule
        \textbf{Method} & \textbf{Caltech101} & \textbf{Food101} & \textbf{SUN397} & \textbf{FGVC Aircraft} & \textbf{EuroSAT} \\
        \midrule
        CLIP & 93.35 & 83.65 & 62.59 & 23.67 & 42.01 \\
        \rowcolor{gray!20} CPE-OT & 95.67$\pm$0.10 & 86.04$\pm$0.04 & 70.56$\pm$0.09 & 32.64$\pm$0.05 & 62.96$\pm$0.64 \\
        \bottomrule
        \end{tabular}
    \end{minipage}
    \vfill
    \begin{minipage}{\textwidth}
        \centering
        \setlength{\tabcolsep}{2pt}
        \begin{tabular}{>{\centering\arraybackslash}p{1.8cm}|*{5}{>{\centering\arraybackslash}p{2.2cm}}}
        \toprule
        \textbf{Method} & \textbf{ImageNet-1k} & \textbf{ImageNet-A} & \textbf{ImageNet-V2} & \textbf{ImageNet-R} & \textbf{ImageNet-K} \\
        \midrule
        CLIP & 66.74 & 47.74 & 60.75 & 73.98 & 46.13 \\
        \rowcolor{gray!20} CPE-OT & 71.70$\pm$0.04 & 60.09$\pm$0.03 & 65.47$\pm$0.02 & 80.46$\pm$0.22 & 52.78$\pm$0.06 \\
        \bottomrule
        \end{tabular}
    \end{minipage}
    \caption{\textbf{Error bar analysis} on 15 image datasets. The mean and the standard deviation of the top-1 accuracy is reported.}
    \label{tab:bar}
\end{table*}

\subsection{License information}
\label{sec:license}
\textbf{Datasets.} Below are the datasets used in this paper that have known license information: 
(1) MIT License: ImageNet-A~\cite{hendrycks2021natural}, ImageNet-V2~\cite{recht2019imagenet}, ImageNet-R~\cite{hendrycks2021many}, ImageNet-Sketch~\cite{wang2019learning}, EuroSAT~\cite{helber2019eurosat}.
(2) CC BY 4.0 License: HMDB51~\cite{kuehne2011hmdb}, Kinetics-600~\cite{carreira2018short}.
(3) CC BY-SA 4.0 License: Oxford Pets~\cite{parkhi2012cats}.

\textbf{Source code.} Source code used in this paper are under the MIT License: CLIP~\cite{radford2021learning}, CoOp~\cite{zhou2022learning}, TPT~\cite{shu2022test}, PLOT~\cite{chen2022plot}, REAL~\cite{parashar2024neglected}.

%-------------------------------------------------------------------------
\section{Limitations}
\label{sec:limit}
Despite the promising results, our method has several limitations. First, the generation of synonyms relies heavily on the quality of LLM outputs, which may produce inconsistent results across different LLM versions or prompting strategies. Second, our activation map-based approach for discriminative region selection may struggle with images containing complex scenes where foreground-background separation is ambiguous. Third, the improvement in performance comes with a slight increase in computational overhead during inference, which may affect real-time applications.

\end{document}